\documentclass[10pt,twocolumn,letterpaper]{article}

\usepackage{cvpr}              

\definecolor{cvprblue}{rgb}{0.21,0.49,0.74}
\usepackage[pagebackref,breaklinks,colorlinks,allcolors=cvprblue]{hyperref}

\def\paperID{*****} 
\def\confName{CVPR}
\def\confYear{2025}

\title{Emergence of Text Readability in Vision Language Models}

\author{First Author\\
Institution1\\
Institution1 address\\
{\tt\small firstauthor@i1.org}
\and
Second Author\\
Institution2\\
First line of institution2 address\\
{\tt\small secondauthor@i2.org}
}

\author{Jaeyoo Park$^{1}\footnotemark[1]$ \quad Sanghyuk Chun$^3$ \quad Wonjae Kim${^3}$\footnotemark[2] \quad Sangdoo Yun$^3$ \quad Bohyung Han$^{1, 2}$ \\ \\
Computer Vision Laboratory, ECE$^1$ \& IPAI$^{2}$, Seoul National University\\
Naver AI Lab$^{3}$\\
{\tt\small \{bellos1203, bhhan\}@snu.ac.kr} \quad {\tt\small \{sanghyuk.c, wonjae.kim, 
sangdoo.yun\}@navercorp.com} 
}

\begin{document}
\maketitle

\footnotetext[1]{Works done during an internship at NAVER AI Lab}
\footnotetext[2]{Now at TwelveLabs}

\begin{abstract}
We investigate how the ability to recognize textual content within images emerges during the training of Vision-Language Models (VLMs). 
Our analysis reveals a critical phenomenon: the ability to read textual information in a given image \textbf{(text readability)} emerges abruptly after substantial training iterations, in contrast to semantic content understanding which develops gradually from the early stages of training.
This delayed emergence may reflect how contrastive learning tends to initially prioritize general semantic understanding, with text-specific symbolic processing developing later.
Interestingly, the ability to match images with rendered text develops even slower, indicating a deeper need for semantic integration.
These findings highlight the need for tailored training strategies to accelerate robust text comprehension in VLMs, laying the groundwork for future research on optimizing multimodal learning.
\end{abstract}

\section{Introduction}
\label{sec:intro}

Vision-Language Models (VLMs) have significantly improved their ability to integrate visual and linguistic information~\cite{chen2024internvl,jia2021scaling,kim2021vilt,desai2023meru,chun2025prolip,radford2021learning,park2023multi}, achieving strong performance in diverse tasks. 
Specifically, CLIP~\cite{radford2021learning} and its variants~\cite{desai2023meru,tschannen2023clippo,chun2025prolip} demonstrate the ability to recognize objects, scenes, and texts within images~\cite{chen2024internvl, materzynska2022disentangling,wang2022git}, enabling applications such as scene-text recognition or document understanding~\cite{park2024hierarchical, ye2023ureader}.
This progress highlights VLMs as powerful tools for bridging the gap between images and text, yet it also raises intriguing questions about the underlying mechanisms driving these capabilities.

Despite these advances, the training dynamics behind VLMs' ability to process textual content within images remain underexplored.
Recently, \citet{lin2024parrot} identifies the source of such ability as 
images containing captions
in large-scale web-crawled training data. However, Lin et al. mainly focus on the negative impact of images containing rendered text on semantic understanding of pure visual contents.
This leaves open questions about the learning trajectories of how VLMs develop text readability and whether this development aligns with their semantic understanding.

\begin{figure}[t]
	\centering
        \includegraphics[width=\linewidth]{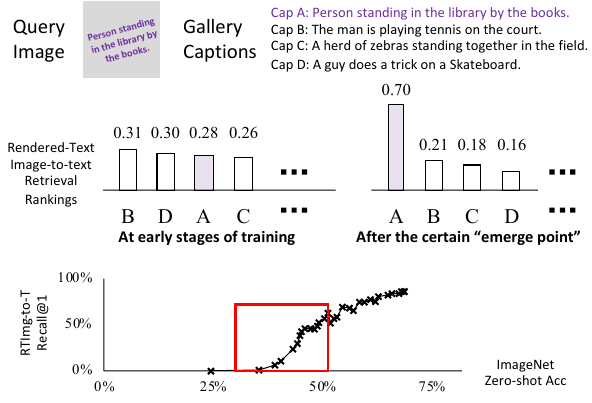}
	\caption{
    Emergence of \textit{text readability} in VLMs during training.
    Our findings show that VLMs can suddenly learn to visually understand rendered text.
    Initially, the model struggles to recognize rendered text inputs (early-stage retrieval rankings, left).
    However, after a certain ``\textit{emerge point}'', its ability to interpret rendered text improves significantly (updated retrieval rankings, right).
    This sharp increase in Rendered-Text Image-to-Text (RTImg-to-T, bottom) occurs only after achieving a certain level of semantic understanding represented by ImageNet zero-shot accuracy.
    }
\label{fig:concepts}
\vspace{-0.4cm}
\end{figure}

In this paper, we focus on the \textit{text readability} in VLMs, \eg, can a VLM ``read'' the texts in an image? 
This can be measured by a retrieval task where a query is an image containing the text and the candidates are the corresponding text (\eg, a rendered text caption such as the ``Person standing in the library by the books'' image and Cap A
in~\cref{fig:concepts}). 
If a model can \textit{visually} understand textual information in an image, the retrieval task can be easily solved.
Our systematic study reveals an interesting insight regarding the text readability in VLMs:
unlike semantic content understanding, which develops gradually from the initial stages, \textit{the ability to read text emerges abruptly after substantial training.} 
This delayed emergence may reflect how optimizing contrastive loss prioritizes semantic learning, while text readability, requiring symbolic understanding, appears with further refinement.
\cref{fig:concepts}~summarizes our findings.

Moreover, we show that the ability to match images with rendered-text images emerges after additional training, lagging behind the development of rendered-text image-to-text \textit{(RTImg-to-T)} matching. 
This underscores the complex interactions of visual and textual features, as the model initially relies on superficial pattern recognition before gradually acquiring deeper semantic integration. 
By highlighting this emergence, our work provides novel insights into the design and interpretability of multimodal systems. 

The rest of the paper is organized as follows.
Section~\ref{sec:related_work} discusses related works on emergent phenomena of foundation models and text readability of VLMs. 
The details of our approach and experimental findings are described in Section~\ref{sec:method}. 
Finally, Section~\ref{sec:conclusion} concludes by summarizing our contributions and proposing future research directions.

\section{Related Work}
\label{sec:related_work}

\subsection{Emergent Abilities of Foundation Models}
Recent studies have explored emergent capabilities in large-scale foundation models, encompassing both large language models (LLMs) and vision-language models (VLMs). 
Research has shown that LLMs exhibit abrupt improvements in tasks like complex arithmetic as their scale increases, suggesting a non-linear progression in ability~\cite{schaeffer2023emergent,snell2024predicting,wei2022emergent}. 
For example, GPT models demonstrate proficiency in zero-shot translation, calculation, and action planning without specific training for these tasks~\cite{achiam2023gpt,brown2020language}. 
Recent findings challenge previous assumptions, suggesting that achieving low pre-training loss—rather than increasing model size or computational power—may be the critical factor behind these emergent abilities across various benchmarks~\cite{du2024understanding}.

In VLMs, a recent study observes that CLIP model acquires the ability to use a red circle as a visual prompt for region-specific focus, which is a skill absent in smaller models, showing how scaling reveals new functionalities~\cite{shtedritski2023does}. 
This ability arises from web-crawled training data with red circles and matching captions. 
Our research reveals that text readability emerges as a distinct VLM capability, similarly shaped by training data with text in images.

\subsection{Text Readability in VLMs}
Recent studies have investigated the text readability of VLMs, uncovering a range of insights into their development. 
\citet{wang2022git} first demonstrated that multimodal large language models, after large-scale pretraining, can read text embedded in images. 
Research into CLIP revealed more intricate mechanisms.
For instance, \citet{materzynska2022disentangling}~show that the image encoder of CLIP can separate word image spelling from natural scene meaning through projection layer finetuning.
Additionally, \citet{gandelsman2023interpreting}~identify specific attention heads specialized in recognizing letter patterns. 
However, \citet{lin2024parrot} highlight a limitation known as ``parrot bias," where models merely learn to recognize the patterns of visible text in images, potentially missing deeper visual understanding. 
Our study explores the temporal dynamics of text-reading development in VLMs, examining its emergence during training to offer a fresh perspective on their evolution.

\section{Method and Experimental Results}
\label{sec:method}

\begin{figure}[t]
	\centering
	\includegraphics[width=\linewidth]{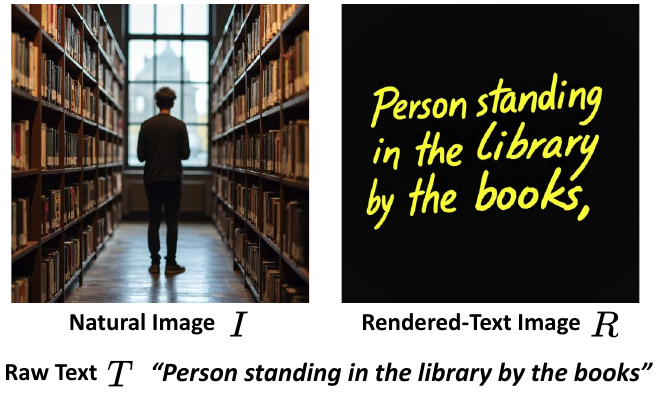}
	\caption{$(T, I, R)$ triplet example generated by FLUX.1 dev using the NoCaps dataset.}
\label{fig:flux_imgs}
\vspace{-0.3cm}
\end{figure}

\begin{figure*}[t!]
	\centering
	\includegraphics[width=0.95\linewidth]{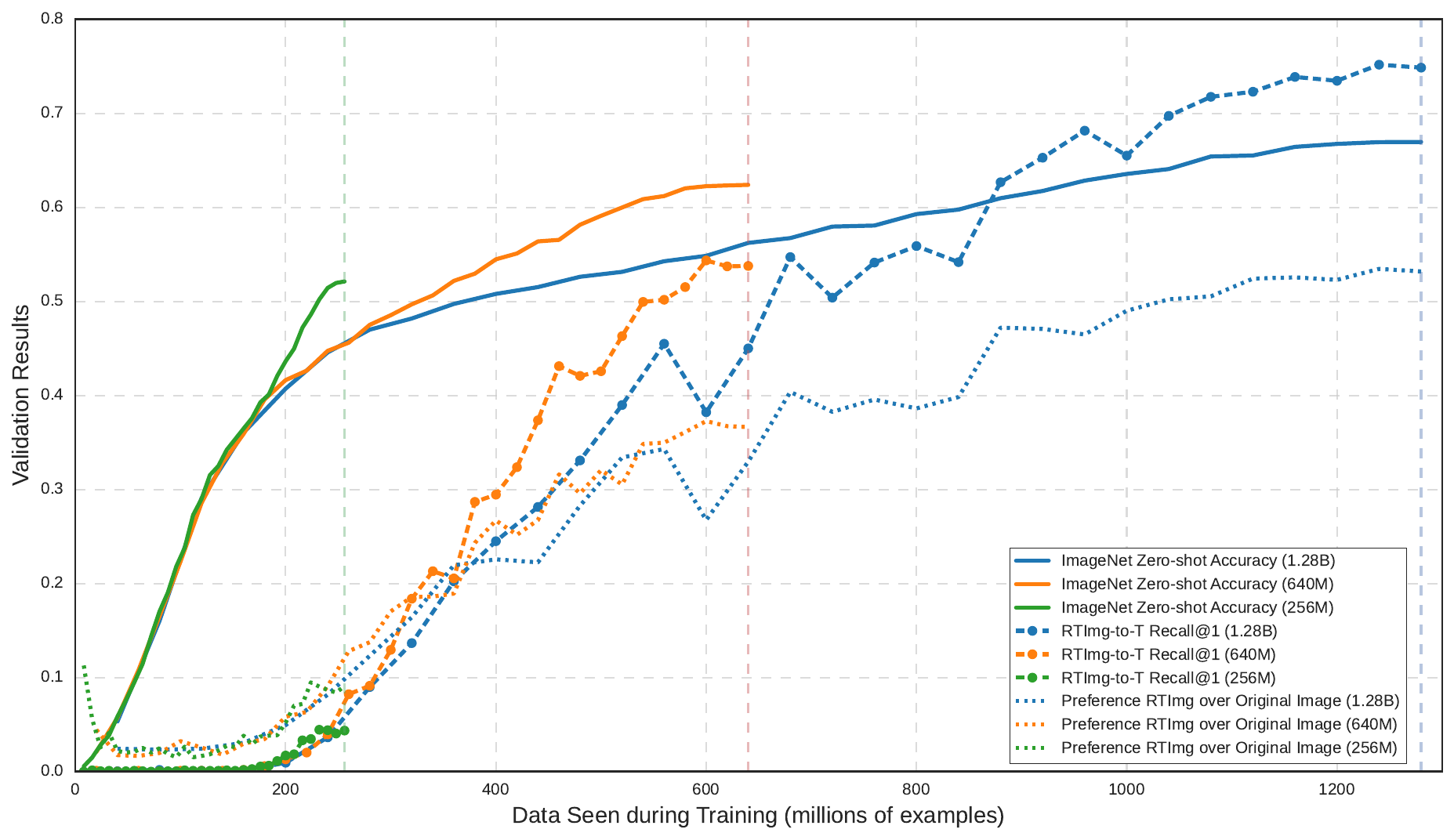}
	\caption{Validation performance of ViT-B/16-based CLIP models trained with three different sample sizes (256M, 640M, and 1.28B) as a function of data seen during training.
    Note that ``RTImg" indicates ``Rendered-Text Image".
    }
\label{fig:seen_sample}
\vspace{-0.3cm}
\end{figure*}

\subsection{Dataset Construction}
\label{subsec:dataset_construction}
Our goal is to systematically investigate the emergence of text readability in VLMs. 
To achieve this, we require triplets consisting of (a) a text description $T$, (b) its corresponding natural image $I$, and (c) an image $R$ with the visually rendered text of $T$ -- an example is shown in~\cref{fig:flux_imgs}.
In this subsection, we introduce how we construct these $(T,I,R)$ triplets.
Our comprehensive evaluation dataset comprises 3,000 high-quality triplets.

We generated these samples by applying a diffusion-based text-to-image pipeline to create images based on captions from the NoCaps dataset~\cite{agrawal2019nocaps}, which provides a diverse corpus of realistic image-caption pairs spanning multiple domains.
For the diffusion model, we employed FLUX.1 dev~\cite{flux2024}, a recent diffusion model recognized for its exceptional text rendering fidelity. 
To ensure dataset diversity and robustness, we varied multiple rendering parameters, including font style, text color, background color, and spatial positioning of the text elements. 
This parametric variation allows for a more thorough evaluation of VLMs' text comprehension ability across different visual contexts.

During dataset curation, we applied a rigorous quality control protocol to filter out samples where text was improperly rendered or became illegible due to visual artifacts or poor contrast. 
This meticulous filtering process ensured that only high-fidelity examples remained in the final dataset, resulting in 3,000 validated triplets suitable for comprehensive analysis.

\subsection{Evaluation Metrics and Protocol}
\label{subsec:evaluation_metrics}
We next explain how we quantitatively measure the text readability in VLMs.
Our primary analysis uses a rendered-text image-to-text retrieval metric (RTImg-to-T Recall@1), which assesses the model's ability to associate rendered-text images with their corresponding raw texts. 
Models with strong text recognition capabilities should achieve high retrieval performance on this task.
Without this ability, its performance should be near random.
Additionally, to assess deeper semantic understanding of rendered text, we employ a rendered-text image-to-image retrieval metric (RTImg-to-I Recall@1), detailed in Section~\ref{sec:semantic}.
We also established a ``preference'' criterion, evaluating if the models assign higher similarity scores for rendered-text images over natural images when matched with the raw text.

We measure the rendered-text image-to-text retrieval metric and preference on ViT-B/16 CLIP models trained on the DataComp-1B dataset~\cite{gadre2023datacomp}, a large-scale collection of 1.4 billion image-text pairs. 
We trained the models with various amounts of seen samples and measured the evaluation metrics during training.
Notably, DataComp-1B likely contains numerous image-caption pairs where captions are rendered directly within the images.
This is primarily because it is constructed from web-crawled data, which naturally includes many rendered texts.
Furthermore, as noted by~\citet{lin2024parrot}, the DataComp filtering process may not effectively exclude ``parroting captions''—captions that merely replicate text embedded in the visual content.
Therefore, CLIP models trained on DataComp-1B are expected to eventually develop text readability. 
Our primary goal is to investigate \textit{when} this ability emerges during training.

To investigate the impact of scale on model performance, our experiments utilize three distinct training dataset sizes (256M, 640M, and 1.28B seen samples).
For each run, we checkpoint 32 intermediate weights with uniform intervals to measure the text readability during training.
Each model is trained on DataComp-1B with batch size of 16,384 using the AdamW optimizer \cite{adamw} by setting $\beta_1=0.9$ and $\beta_2=0.95$. 
During training, a random scaling (40\% to 100\%) and random color jittering are applied for data augmentation.

\subsection{Emergence of Text Readability}
To investigate how text readability develops during training, we analyzed the performance of ViT-B/16 CLIP models trained on DataComp-1B with varying sample sizes (256M, 640M, and 1.28B), as shown in~\cref{fig:seen_sample}. 
The figure plots ImageNet zero-shot classification accuracies (solid lines) and text readabilities (dashed lines), alongside preferences (dotted lines) of each model. 
Notably, text readability and the preference for rendered text images emerge consistently around 200 million samples, irrespective of the learning schedule or sample size. 
Despite the abundant presence of rendered captions in the training dataset and consistent improvements in ImageNet zero-shot accuracy, text readability emerges later in training. 
This delayed emergence may reflect how contrastive loss optimization tends to prioritize semantic learning, while symbolic representation learning—linked to text readability—appears to develop with further loss optimization.

\begin{figure}[t]
	\centering
	\includegraphics[width=\linewidth]{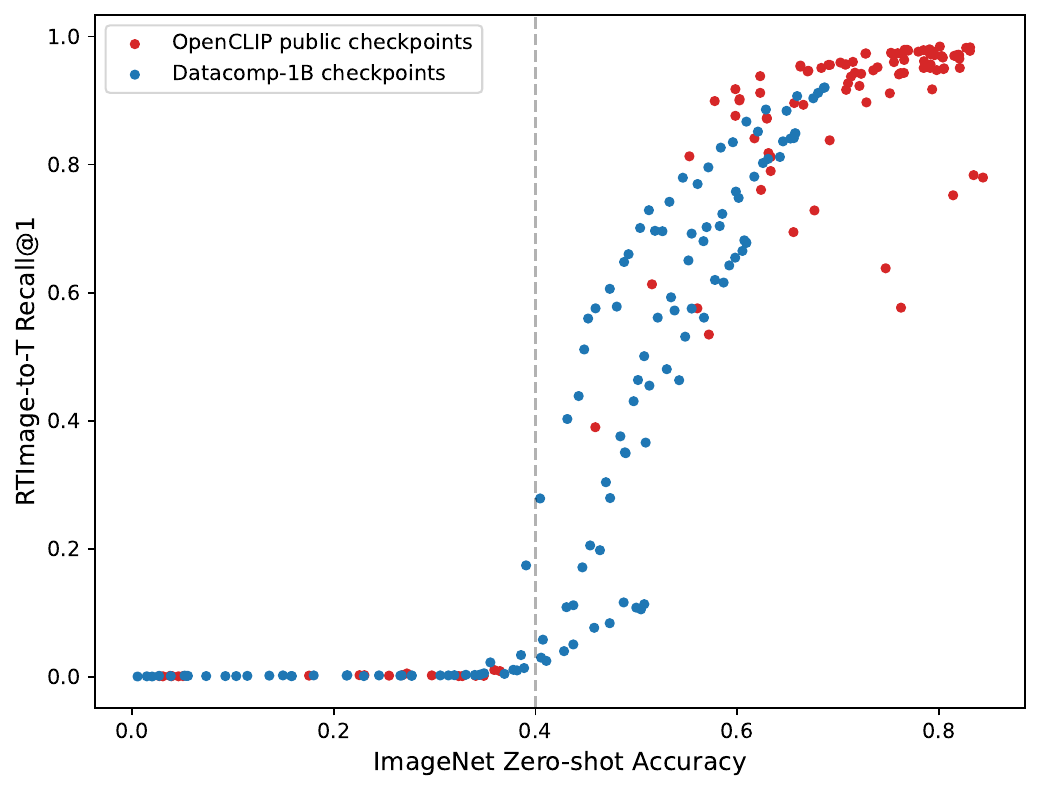}
	\caption{ImageNet zero-shot classification accuracy vs text readability for CLIP models. OpenCLIP public checkpoints (red) and Datacomp-1B checkpoints (blue) are distinguished by color. The plot highlights a non-linear correlation, with text readability emerging around 0.4 ImageNet zero-shot accuracy.}
\label{fig:imagenet}
\vspace{-0.4cm}
\end{figure}

Building on our previous analysis, we further investigated the relationship between text readability and semantic understanding by evaluating 114 publicly available CLIP models from the OpenCLIP repository~\cite{ilharco_gabriel_2021_5143773}. 
For these public models, alongside our trained models, we assessed text readability and its correlation with semantic understanding. 
Semantic understanding was measured using ImageNet zero-shot classification accuracy~\cite{russakovsky2015imagenet}, a standard metric for CLIP performance. 
Note that the public OpenCLIP weights include diverse architectures (\eg, ViT-G~\cite{dosovitskiy2020image}, ConvNext~\cite{liu2022convnet}, SigLIP~\cite{zhai2023sigmoid}) and pretraining datasets (\eg, LAION-2B~\cite{schuhmann2022laion}, WebLi~\cite{chen2022pali}, DFN~\cite{fang2023data}), allowing for a broader examination.

\cref{fig:imagenet}~illustrates this relationship, distinguishing OpenCLIP models (red) from the Datacomp-1B-trained checkpoints (blue) by color.
From the results, we observed a non-linear correlation between ImageNet zero-shot accuracy and text readability, with readability emerging abruptly around an accuracy of 0.4 regardless of the model type. 
This pattern holds consistently for Datacomp-1B-trained models across different seen samples during training (256M, 640M, 1.28B), indicating that the abrupt emergence of text readability is generalized to variations in training scale.
This observation aligns with our hypothesis that the contrastive loss optimization process prioritizes semantic content understanding over symbolic representation learning, though both capabilities eventually develop with sufficient training.
\vspace{-0.4cm}

\begin{figure*}[t!]
	\centering
        \begin{subfigure}[c]{0.49\linewidth}
            \centering
            \includegraphics[width=\linewidth]{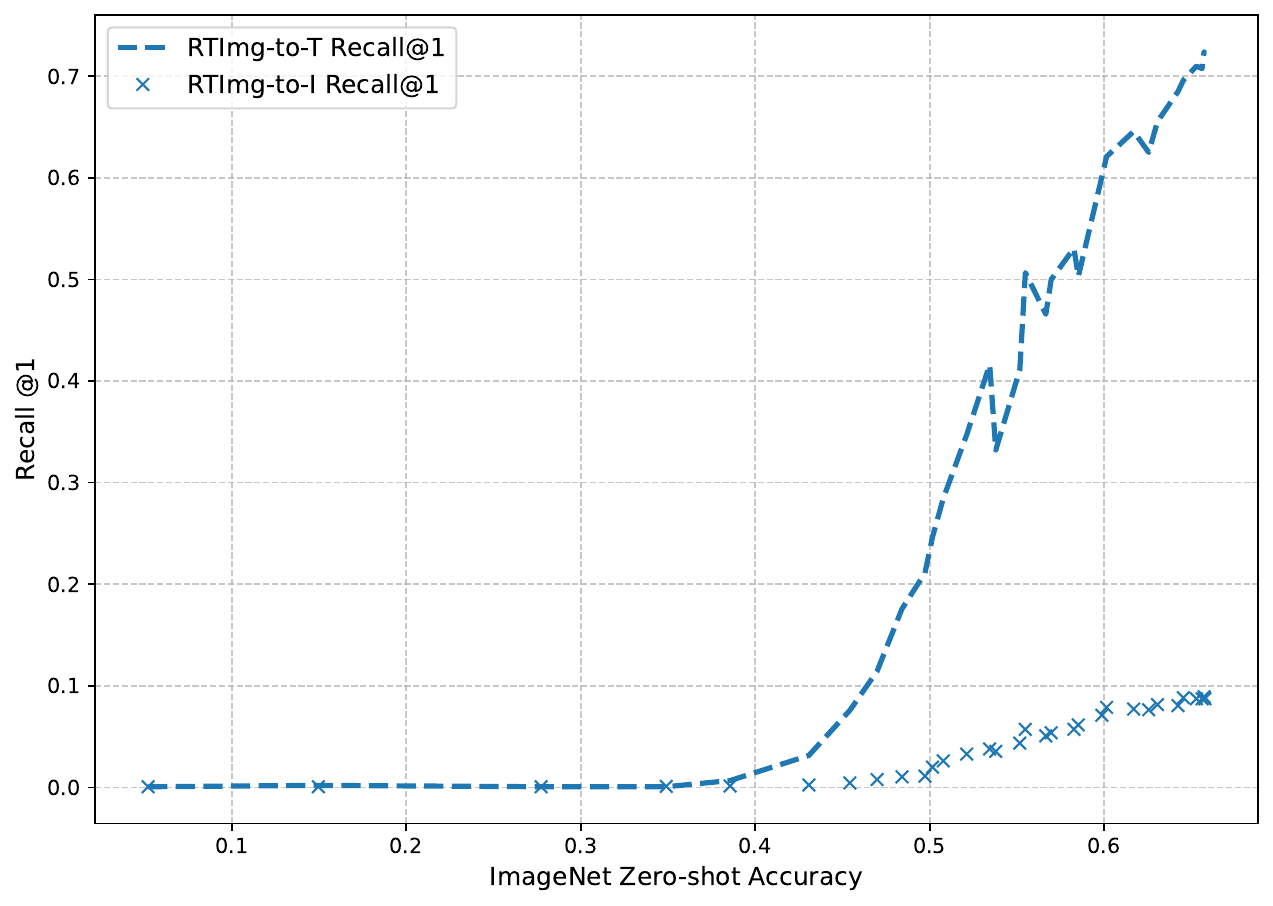}
        \end{subfigure}
	\begin{subfigure}[c]{0.49\linewidth}
            \centering
            \includegraphics[width=\linewidth]{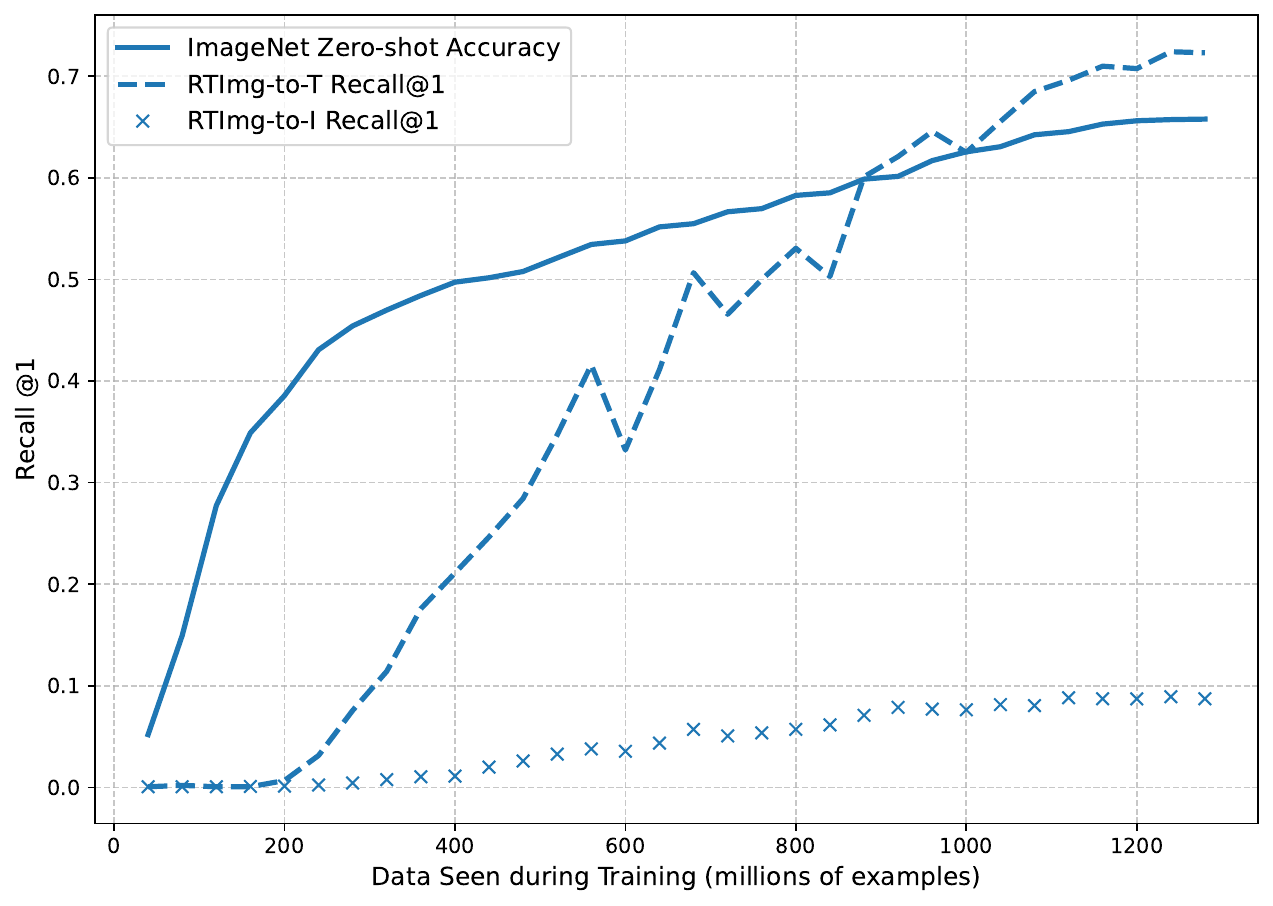}
        \end{subfigure}
\caption{Comparison of CLIP models’ matching performance between rendered-text images and text versus rendered-text images and natural images across learning trajectory. 
The graph shows lower performance in RTImg-to-I matching, indicating reliance on pattern recognition, and delayed emergence of this ability compared to RTImg-to-T matching, reflecting the challenges of contrastive learning in aligning visually processed inputs.}
\label{fig:image_rendered}
\end{figure*}

\begin{figure*}[t!]
	\centering
        \begin{subfigure}[c]{0.49\linewidth}
            \centering
            \includegraphics[width=\linewidth]{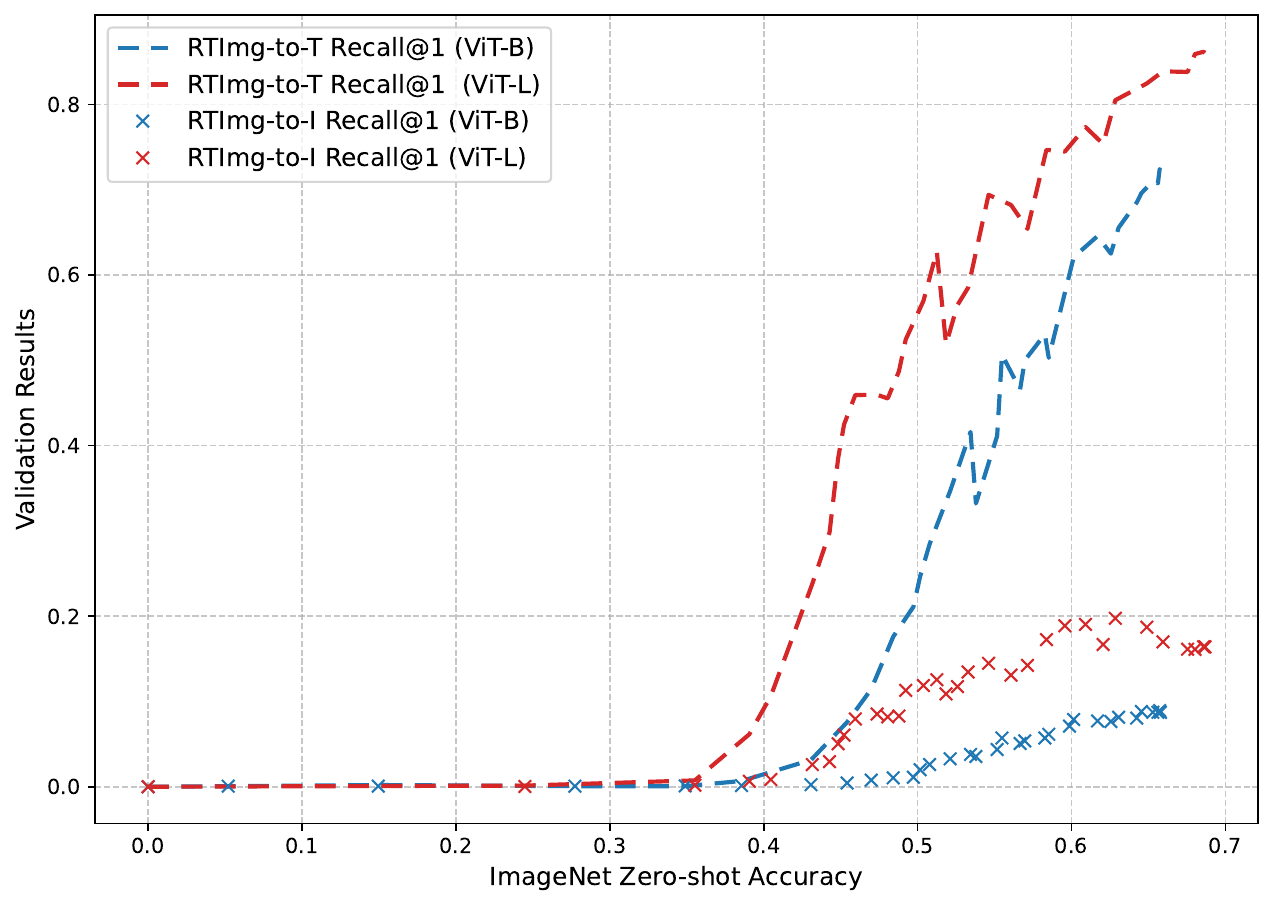}
        \end{subfigure}
	\begin{subfigure}[c]{0.49\linewidth}
            \centering
            \includegraphics[width=\linewidth]{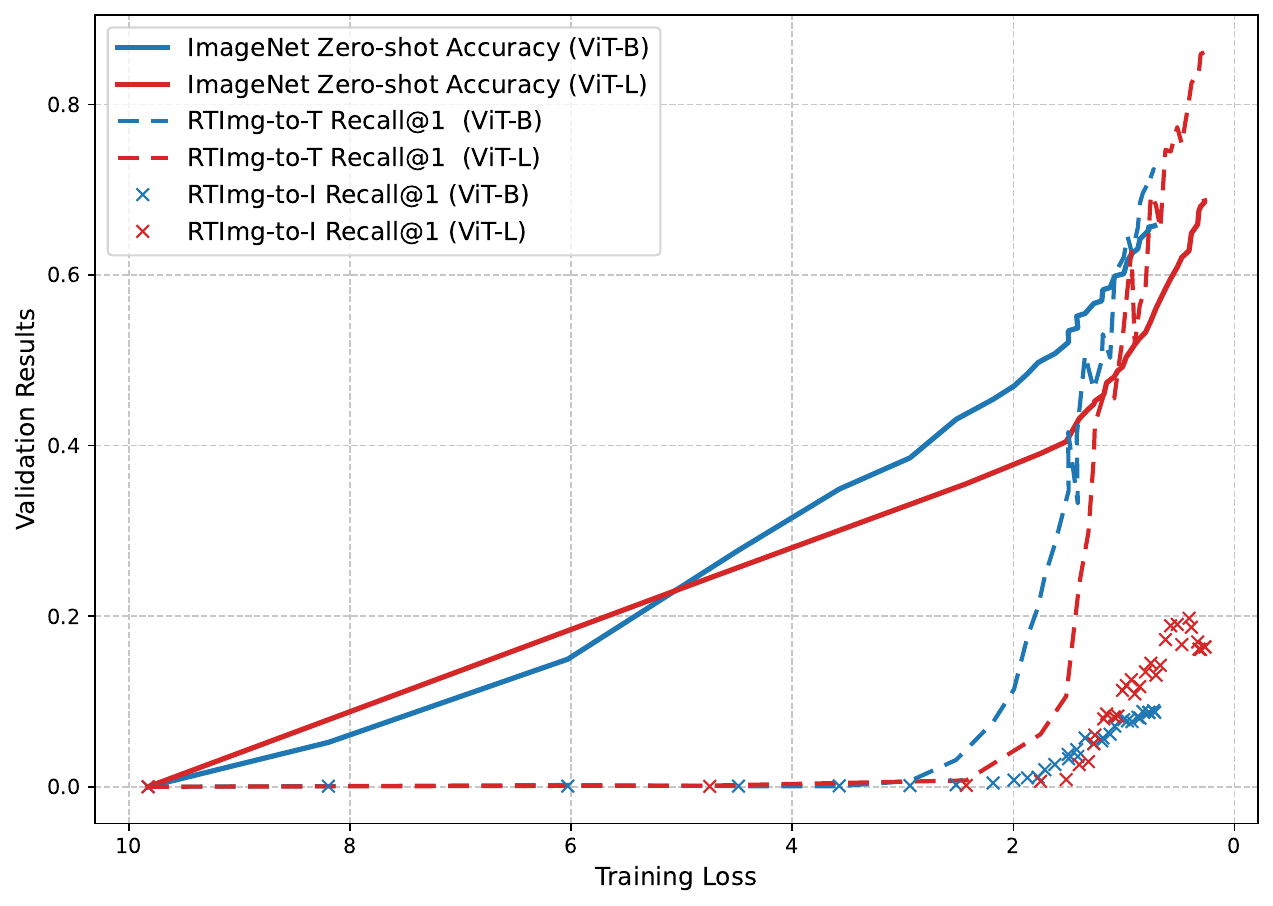}
        \end{subfigure}
\caption{Comparison of text readability and semantic understanding of rendered-text images for ViT-B/16 and ViT-L/16 models, plotted against ImageNet zero shot accuracy and training loss.}
\label{fig:model_scale}
\vspace{-0.3cm}
\end{figure*}

\subsection{Semantic Understanding of Rendered Text} 
\label{sec:semantic}
Thus far, our analysis has focused on RTImg-to-T matching, which serves as an indicator of the model’s text readability. 
However, this approach may reflect a na\"ive pattern-matching strategy as \citet{lin2024parrot} claimed, raising the question of whether the model can truly grasp the semantic meaning of rendered text. 
If a model truly understands rendered text and accurately embeds the meaning of the text in its representation space, the visual embedding of rendered text image ($R$), the textual embedding of the original caption ($T$), and the embedding of its corresponding natural image ($I$) should be closely aligned. 
Namely, we expect that not only the RTImg-to-T ability emerges but also the RTImg-to-\textbf{I} ability emerges similarly.

We report the RTImg-to-I Recall@1 in \cref{fig:image_rendered} along with RTImg-to-T Recall@1. 
This comparison reveals two key findings.
First, RTImg-to-I recall is significantly lower compared to RTImg-to-T recall (\eg, below 10\% vs. above 70\%). 
This finding may suggest that the emergence of the RTImg-to-T ability could be a reliance on superficial pattern recognition rather than deep semantic understanding.

Second, the RTImg-to-I matching ability emerges later than the emergence of the RTImg-to-T matching ability.
We presume that this delay is likely because contrastive learning primarily involves direct comparisons between images and texts, rather than between rendered-text images and images—despite both being visual inputs. 
However, this RTImg-to-I capability gradually improves with sufficient training (e.g., achieving around 10\% Recall@1 with over 1B seen samples).
Furthermore, we also observe that if we use a stronger model (\eg, ViT-L/16), the RTImg-to-I recall@1 is substantially improved (\eg, 20\%). 
More details are in the next subsection.

Overall, we presume that the emergence of RTImg-to-T Recall@1 would not solely originate from the truly semantic and visual understanding of rendered text, but it also could rely on a simple pattern matching of rendered text (\eg, detecting a specific word, rather than understanding the whole caption semantically).
Developing a training strategy to achieve both matching abilities (RTImg-to-T and RTImg-to-I) will be an interesting future research direction.

\subsection{Scaling Model Capacity with ViT-L/16}
We extended our analysis to a larger ViT-L/16 backbone, training a CLIP model on the DataComp-1B dataset. 
This ViT-L/16 model used the same training configuration as its ViT-B/16 counterpart, trained with 1.28B seen samples.
Following the methodology outlined by~\citet{du2024understanding}, we also plotted the text readability and semantic understanding of rendered-text images for both ViT-B/16 and ViT-L/16 against training loss in~\cref{fig:model_scale}. 
The results indicate that the pattern previously observed in the smaller model—where text readability and semantic understanding of rendered text emerge after the model grasps semantic understanding of images—also holds for the larger ViT-L/16; and furthermore, a larger model exhibits stronger text readability. 
\vspace{-0.1cm}

\section{Conclusion}
\label{sec:conclusion}

In this paper, we have explored the emergence of textual content recognition in Vision-Language Models (VLMs), uncovering significant patterns in their training dynamics. 
Our central finding is that the ability to read text (RTImg-to-T) arises abruptly after extensive training, a stark contrast to the steady development of general semantic understanding.
This delay likely reflects how contrastive learning prioritizes broader visual-semantic features before symbolic capacities such as text recognition.
Models seem to first prioritize general visual-semantic features before developing symbolic capacities like text recognition. 
The difficulty of tasks like RTImg-to-I further indicates that deeply integrating rendered text visually is a more advanced capability.
These insights emphasize the importance of tailoring training approaches to foster not just text recognition but genuine understanding in VLMs. 
We hope this study sets the stage for future investigations into refining multimodal learning strategies.

{
    \small
    \bibliographystyle{ieeenat_fullname}
    \bibliography{main}
}


\end{document}